\documentclass{article}
\usepackage{amssymb,amsmath}
\usepackage{graphicx}
\usepackage{url}
\usepackage{xcolor}
\usepackage{authblk}
\usepackage{geometry}

\newgeometry{vmargin={25mm}, hmargin={25mm,20mm}}

\def\dsp{\displaystyle}

\def\RR{\mathbb{R}}

\title{The coupling effect of Lipschitz regularization\\ in deep neural networks}

\author[$\dagger$]{Nicolas Couellan}

\affil[$\dagger$]{\small{ENAC, Universit\'{e} de Toulouse, 7 Avenue Edouard Belin, 31400 Toulouse, France} 
E-mail:\texttt{nicolas.couellan@recherche.enac.fr}}

\date{}

\begin{document}
\maketitle

\abstract{
We investigate robustness of deep feed-forward neural networks when input data are subject to random uncertainties. More specifically, we consider regularization of the network by its Lipschitz constant and emphasize its role. We highlight the fact that this regularization is not only a way to control the magnitude of the weights but has also a coupling effect on the network weights accross the layers. We claim and show evidence on a dataset that this coupling effect brings a tradeoff between robustness and expressiveness of the network. This suggests that Lipschitz regularization should be carefully implemented so as to maintain coupling accross layers.  
}
\section{Introduction}

With the increasing interest in deep neural networks, a lot of research work has been focusing lately on their sensitivity to input perturbations \cite{Fawzi2018,Shaham2018,Wong2018,Raghunathan2018}. Most of these investigations have highlighted their weakness to handle adversarial attacks and the addressed means to increase their robustness. Adversarial attacks are real threats that could slow down or eventually stop the development and applications of these tools wherever robustness guarantees are needed. If input uncertainty is not adversarial but simply generated by the context and environment of the specific application, deep networks do not show better behavior in terms of robustness.  If no immunization mechanism is used, their generalization performance can greatly suffer. Actually, most of the techniques that are used to handle adversarial attacks can be also applied in this context. These techniques can mostly be classified into two categories: robust optimization techniques that consider an adversarial loss \cite{Shaham2018,Wong2018} and regularization techniques that penalize noise expansion throughout the network \cite{Gouk2018,Oberman2018,Yoshida2018,Finlay2018,Tsuzuku2018,Scaman2018}.

In this article, we address the second type of techniques where robustness can be achieved by ensuring that the Lipschitz constant of the network remains small. The network can be seen as a mapping between inputs and outputs. Its robustness to input uncertainties can be controlled by how much the mapping output expands the inputs. In the case network with Lipschitz continuous activations, this is equivalent to controlling the Lipschitz constant of the whole network mapping.

Expressiveness is another important property of the neural network. It defines the ability of the network to represente highly complex functions. It is achieved by depth \cite{Eldan2016,Raghu2017}. Of course, such ability is also to be balanced with its generalization power in order to avoid overfitting during training. On one hand, if the weights of the networks are free to grow too high, the generalization power will be low. On the other hand, if the weights are overly restricted, the expressiveness of the network will be low. Usually, this tradeoff is controlled by constructing a loss that accounts for both training error and generalization error, the so-called regularized empirical risk functional. A parallel has been established between robustness and regularization \cite{xu}. Actually in the case of support vector machines, it has been shown that both are equivalent. The work on deep regularized networks we have mentionned above suggests that this is also true for neural networks. The idea we are developing here is a contribution along this line. 

We argue that Lipschitz regularization does not only restrict the weights magnitude but it also implements a coupling mechanism accross layers, allowing some weights to grow high while others are allowed to vanish. This happends accross layers, meaning that the Lipschitz constant can remain small with some large weights values in one layer while some other weight in other layers counterbalance this growth by remaining small. On the contrary, if all weights are restricted uniformly accross the network, even though network depth is large, the expressiveness of the network may be inhibited. Through a numerical experiment on a popular dataset, we show evidence of this phenomenon and draw some conclusions and recommmandations about software implementation of Lipschitz regularization.    

In Section~\ref{lipschitz}, we introduce the relationship between network robustness and Lipschitz regularization and discuss the coupling mechanism taking place. Section \ref{experiment} is an illustration of the phenomenon through a numerical experiment and Section \ref{conclusion} concludes the article.\\

\section{Neural Network Robustness as a Lipschitz constant regularization problem}\label{lipschitz}
 
\subsection{Propagating additive noise through the network}

Consider feed-forward fully connected neural networks that we represent as a successive composition of linear weighted combination of functions such that $x^l=f^l(W^lx^{l-1}+b^l)$ for $l=1,\ldots,L$, where $x^{l-1}\in \RR^{n_{l-1}}$ is the input of the $l$-th layer, the function $f^l$ is the $L_f$-Lipschitz continuous activation function at layer $l$, and $W^l\in \RR^{n_l\times n_{l-1}}$ and $b^l\in \RR^{n_l}$ are the weight matrix and bias vector between layer $l-1$ and $l$ that define our model parameter $\theta=\{W^l,b^l\}_{l=1}^{L}$ that we want to estimate during training. The network can be seen as the mapping $g_\theta:x^0\rightarrow g_\theta(x^0)=x^L$. The training phase of the network can be written as the minimization of the empirical loss $\mathcal{L}(x,y,\theta)=\frac{1}{n}\sum_{i=1}^{n}l_\theta(g_\theta(x_i),y_i)$ where $l_\theta$ is a measure of discrepancy between the network output and the desired output.

Assume now that the input sample $x_i$ is corrupted by some bounded additive noise $\delta_i$ such that $\forall i\in\{1,\ldots,n\}, \quad \|\delta_i\|\leq \Gamma_i$ for some positive constant $\Gamma_i$. We define $\tilde{x}_i^l=x_i^l+\delta_i^l$ as the noisy observation of $x_i^l$ that we obtain after propagating a noisy input through layer $l$. We can write $\|\delta_i^{l}\|=\|\tilde{x}_i^{l}-x_i^{l}\|=\|f(W^l\tilde{x}_i^{l-1})-f(W^l x_i^{l-1})\|$ that we can upper bound by $L_f\|W^l(\tilde{x}_i^{l-1}-x_i^{l-1})\|=L_f\|W^l\delta_i^{l-1}\|$ since $f$ is $L_f$-Lipschitz continuous. Therefore, the network mapping $g_\theta$ is $L_{g_\theta}$-Lipschitz continuous and the propagation of the input noise throughout the whole network, leads to an output noise that satisfies the following:
$$
\|\delta_i^L\|\leq L_f^L \|W^1\| \times \|W^2\| \times \ldots \times \|W^L\| \Gamma_i
$$
where $\|W^l\|$ denotes the operator norm:
$$
\dsp\|W^l\|=\sup_{x\in\RR^{n_l*}}\frac{\|W^l x\|}{\|x\|} 
$$
and the quantity $\hat{L}_{g_\theta}=L_f^L \|W^1\| \times \|W^2\| \times \ldots \times \|W^L\|$ is actually an upper bound of $L_{g_\theta}$. 

\subsection{Network noise contraction by controlling its Lipschitz constant}
In this section, for simplicity, but without loss of generality, we will consider $1$-Lipschitz activation functions, meaning that $L_f=1$ (this is for example the case of ReLu functions).

We have just seen that the quantity $\hat{L}_{g_\theta}(\theta)=\prod_{l=1}^{L}\|W^l\|$ says how much the input noise will be expanded after propagation through the network. Therefore, if during the training process, we ensure that this quantity remains small, we also ensure that input uncertainties will not be expanded by the successive neurons layers. There are two ways to control this quantity during training:\\

\noindent\textit{Constrained optimization}: The idea is to solve the following empirical risk minimization problem:
$$
\min_{\theta=(W^l,b^l)_{l=1}^L} \mathcal{L}(x,y,\theta) \mbox{ st } \prod_{l=1}^{L}\|W^l\|\leq L_{max}
$$
where $L_{max}$ is positive a parameter. The difficulty with this approach is the nonlinearity of the constraint. One would like to use aprojected stochastic gradient method to solve the training problem. However projecting onto this constraint is a difficult problem. To do so, in \cite{Gouk2018,Yoshida2018}, the authors have proposed to compute and restrict $\|W^l\|$ layer by layer instead of restricting the whole product. Restricting the norm of the weights layer by layer is actually very different from restricting the product of their norms. The layer by layer process isolates the tuning of weights while if the whole product is considered, some layers may be privileged against other. We will see in the next section how this can affect, for some dataset the expressiveness of the network.\\ 

\noindent\textit{Lipschitz regularization}: The alternative is to introduce a regularization term in the loss as follows:
$$
\min_{\theta=(W^l,b^l)_{l=1}^L} \frac{1}{\lambda}\mathcal{L}(x,y,\theta) + \prod_{l=1}^{L}\|W^l\|
$$
where $\lambda$ is a positive parameter. There are no projection involved, the regularization acts through the addition of a correction term in the gradient of the loss so as to ensure a low value of the Lipschitz constant. The gradient of the regularized loss, denoted $\mathcal{L}_r$ can be written as:
$$
\nabla_\theta \mathcal{L}_r(x,y,\theta)=\frac{1}{\lambda}\nabla_\theta\mathcal{L}(x,y,\theta)+\nabla_\theta \hat{L}_{g_\theta}(\theta)
$$
and, as mentioned above, depends on the complete cross-layer structure via $\nabla_\theta \hat{L}_{g_\theta}(\theta)$. To further emphasize this coupling effect, under the assumption that $L_f=1$, we can rewrite $\hat{L}_{g_\theta}(\theta)$ as
$$
\hat{L}_{g_\theta}(\theta)=\sqrt{\lambda_{max}({W^1}^{\top}W^1)}\times\ldots\times\sqrt{\lambda_{max}({W^L}^{\top}W^L})
$$
where $\lambda_{max}(A)$ denotes the largest eigenvalue of matrix $A$. We see in this last expression that, if we could rotate the matrix ${W^l}^{\top}W^l$ at each layer $l$ such that the principal axes are aligned with its eigenvectors, the upper bound $\hat{L}_{g_\theta}(\theta)$ would only depend on the product layer by layer of the largest weight length along these axes. The upper bound could then be seen as a layer by layer product of weight "size". This also means that if at one layer, weights are small, there is room for increase at another layer as long at the whole product remains small. In this sense, we say that the weights have more degrees of freedom than when they are restricted at each layer independently.
\begin{figure}
\begin{center}
\begin{tabular}{cc}
\includegraphics[scale=0.4]{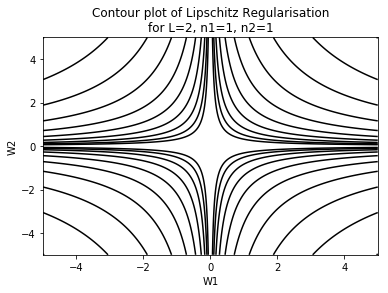} 
\includegraphics[scale=0.4]{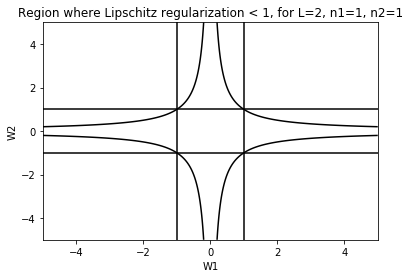} 
\end{tabular}
\end{center}
\caption{Contours of $\hat{L}_{g_\theta}(\theta)$ (left) when $\theta\in\RR^2$ and regions where $\hat{L}_{g_\theta}(\theta)<1$ and $|W^1|<1,|W^2|<1$ with $\theta=(W^1,W^2)$ (right)}\label{lipreg}
\end{figure}
This is also illustrated on a very simple example in Figure~\ref{lipreg}(right). We take the very simple case of a one hidden layer neural network with one input, one hidden and one output neuron, meaning that the parameter $\theta=(W^1,W^2)$ belongs to $\RR^2$. In this case, the bound $\hat{L}_{g_\theta}(\theta)$ is equal to $|W^1|\times |W^2|$. The right figure shows the boundary $|W^1|\times |W^2|=1$ and the center square defines the set $\{(W^1,W^2)\in\RR^2:|W^1|\leq 1,|W^2|=1\}$ which is the restriction of each layer weight matrix to 1 independently. It clearly shows that this layer by layer restriction is more conservative than the Lipschitz upper bound that allows some $W^1$ to be large when $W^2$ is small or the other way around. This is what we refer to as the coupling mechanism accross layers. Figure~\ref{lipreg} only shows a very low dimensional case. If the dimension is large, it easy to understand that, for a specific level of the regularization, the feasible set of the weights for the Lipschitz regularization will be much larger than the feasible set of restricted weights at each layer. We argue and show in the next section that for some specific dataset, this extra feasible volume for the weights enables better expressive power of the network. Note that, computationally, there are several difficulties with this coupling approach:

First, the Lipschitz regularization or more precisely, its upper bound is a non convex function as shown for example on the 2D example of Figure~\ref{lipreg}(left). Minimizing such a function may be difficult and available training algorithms such as stochastic gradient techniques or its variants may get trapped in a local minimum. This is not specific to this case. Non convex regularization techniques have been proven to be effective in other contexts while facing the same difficulty \cite{Wen2018}. However, in practice, the benefit is often confirmed.

The computation of $\nabla_\theta \hat{L}_{g_\theta}(\theta)$ may also be difficult. In practice, it requires the use of numerical differentiation since there is no simple explicit expression of the gradient. In \cite{Yoshida2018}, alternatively, the authors have used a power iteration method to approximate the operator norm. Observe, however, that since the network parameter values must already be stored at any time during the training process, the computation of $\nabla_\theta \hat{L}_{g_\theta}(\theta)$ does not increase storage requirements. 

With respect to these numerical difficulties, please also note that we only emphasize the role of the coupling effect of the Lipschitz regularization and we do not claim to provide efficient techniques to handle it especially on large problems. However, we want to point out that the future development of efficient robust neural network algorithms should preserve the cross layer structure of the regularization. The design of methods that isolate layers are of course computationally interesting but will loose some of the property of Lipschitz regularization and may turn out to be over-conservative in terms of robustness, at least on some datasets, as we will see in the next section.

\section{Experiment}\label{experiment}

To illustrate the coupling effect discussed above, we consider the deep neural network regression task with the Boston dataset \cite{boston}. For this purpose, we use a 3 hidden layers feed-forward fully connected network using ReLu activation functions. The network has 3 hidden layers having each 20 neurons. For training and testing, we use the \texttt{keras} library \cite{keras} under the \texttt{python} \cite{python} environment.
We choose to compare four various training loss formulations and compare, during training, their validation mean average error, their loss values as well as the spectral norm of the weights matrix of each layer. The tested formulations contain all the mean average square loss but include various regularizations or limiting mechanisms on weigths, as follows:
\begin{itemize}
\item[-] (No reg) no regularization
\item[-] (Layer reg) spectral norm regularization at each layer (no coupling)
\item[-] (Lipschitz reg) Lipschitz regularization across layers (with coupling)
\item[-] (MaxNorm) MaxNorm constraint on weights (max value = 10 on weights at each layer) as described in \cite{maxnorm}
\end{itemize}

\begin{figure}
\begin{center}
\begin{tabular}{cc}
\includegraphics[scale=0.4]{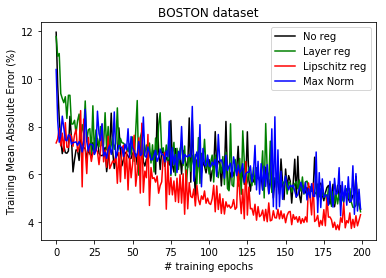}&
\includegraphics[scale=0.4]{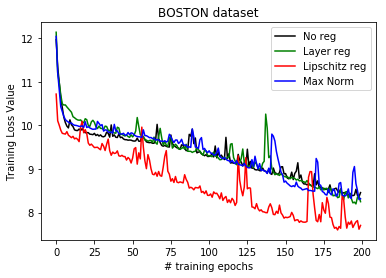}\\
\end{tabular}
\caption{Training mean average error and training loss profiles (Boston dataset)}\label{trainingprofiles}
\end{center}
\end{figure}

\begin{figure}
\begin{center}
\begin{tabular}{cc}
\includegraphics[scale=0.4]{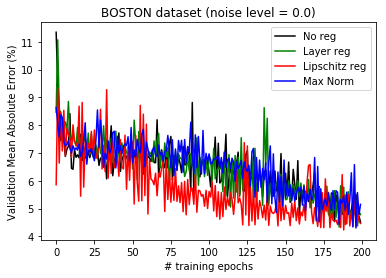}&
\includegraphics[scale=0.4]{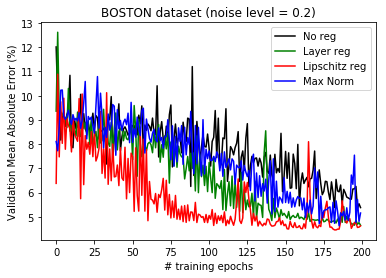}\\
\includegraphics[scale=0.4]{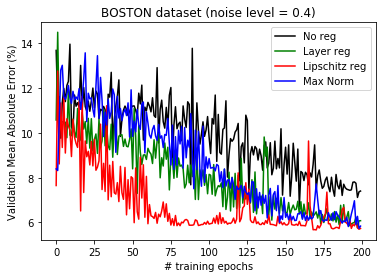}&
\includegraphics[scale=0.4]{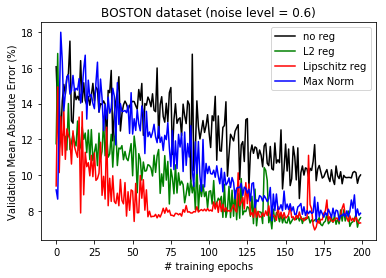}
\end{tabular}
\caption{Mean absolute validation error profiles (Boston dataset)}\label{maeprofiles}
\end{center}
\end{figure}

The \texttt{ADAM} optimization algorithm \cite{adam} is used for training. To implement the "Layer reg" and "Lipschitz reg", a custom regularization and a custom loss were created respectively in \texttt{keras}. The training procedure was set to $200$ epochs with a batch size of $50$. The regularization parameter was selected by grid search. The training is carried out on a fraction of $4/5$ of the entire dataset without perturbing the input data. However, we validate the various formulations with several levels of test uncertainties to evaluate the robustness of each formulation on the remaining fraction of the data. The noisy test inputs are generated as follows:

$$
\tilde{x_i}=x_i + \delta_i \quad \forall i\in T
$$
where $T$ is the test set, $x_i$ is a nominal input set aside before training from the \texttt{Boston} dataset and $\delta_i$ is an additive uncertainty such that $\delta_i = \eta (x_i^{max}-x_i^{max})u_i$ where $u_i\sim U([0,1])$ (the uniform distribution on the interval $[0,1]$), $\eta\in\{0,0.2,0.4,0.6\}$ is the noise level and $x_i^{min}$ and $x_i^{max}$ are the vectors of minimum values and maximum values for each input features. Figure~\ref{trainingprofiles} provides the training mean average error and loss profiles during training while Figure~\ref{maeprofiles} gives the mean average validation error for the various noise levels $\eta$.

\begin{table}
\begin{center}
\caption{Spectral norm of layer weights and network Lipschitz constant upper bound}\label{tablelip}
\begin{tabular}{c|cccc}
Noise level & no reg & Layer reg & Lipschitz reg & Max Norm \\
\hline
$\|W_1\|$ & 2.828 & 2.508 & 3.464 & 2.383 \\
$\|W_2\|$ & 1.954 & 1.625 & 1.791 & 1.883 \\
$\|W_3\|$ & 1.825 & 3.315 & 1.983 & 1.707 \\
\hline
$\hat{L}_{g_\theta}(\theta)$       & 10.01 & 13.52 & 12.31  &  7.66       
\end{tabular}
\end{center}
\end{table}

During training, Figure~\ref{trainingprofiles} shows that the Lipschitz regularization achieves better mean absolute error and loss values than the other techniques that achieve all together similar results. One could suspect overfitting of the data during training with Lipschitz regularization but the validation phase on unseen data as shown on Figure~\ref{maeprofiles} actually does not confirm this. The mean absolute validation error achieved by the Lipshitz regularization is better than the other methods. This is confirmed for all levels of uncertainties, meaning that Lipschitz regularization is, for the Boston dataset, the technique that provides the highest level of robustness. More specifically, it is worth noticing that the layer-by-layer Lipschitz regularization as opposed to the Lipschitz regularization accross layer is not performing well here. At the highest noise level $\eta=0.6$, the Lipschitz regularization achieves a low mean absolute validation error twice as fast as the layer-by-layer approach. The MaxNorm approach achieves better results than the non regularized model but is behind the layer-by-layer approach. These results are consistant with the fact that Lipschitz regularization provides good level of accuracy. When looking at Table~\ref{tablelip}, we can observe some significant differences in the spectral norms of layer weights for the various network instances. All formulations tend to emphasize the first layer except the layer-by-layer approach that allocates more weight mass at the last layer. The Lipschitz regularization accross layers tends also to achieve higher weight values than others, which is natural when considering the shape of the regularization as shown in Figure~\ref{lipreg}. The value of the network Lipschitz constant are similar except for the MaxNorm case that tends to restrict more the weights at all layers. This example confirms that the Lipschitz regularization (accross layers) provides more freedom to the weights than other techniques for the same value of the network Lipschitz constant. This confirms our idea that, for some datasets, letting the regularization play a coupling mechanism accross layers helps in finding the best compromise between robustness and expressiveness of the network.  

\section{Conclusions}\label{conclusion}

In this article, we have discussed some properties of Lipschitz regularization used as a robustification method in deep neural networks. Specifically, we have shown that this regularization does not only control the magnitude of the weights but also their relative impact accross the layers. It acts as a coupling mechanism accross layers that allows some weights to grow when other counterbalance this growth in order to control the expansion of noise throughout the network. Most of Lipschitz regularization implementations we are aware of actually isolate the layers and do not benefit from the coupling mechanism. We believe that this knowledge should be useful in the future and help designing robust neural network training that achieves also good expressiveness properties.

\bibliographystyle{abbrv}
\bibliography{refs}

\end{document}